\documentclass[runningheads]{llncs}

\usepackage{eccv}

\usepackage{eccvabbrv}

\usepackage{graphicx}
\usepackage{booktabs}
\usepackage{times}
\usepackage{epsfig}
\usepackage{amsmath}
\usepackage{amssymb}
\usepackage[ruled,linesnumbered]{algorithm2e}
\usepackage{multirow}
\usepackage{multicol}
\usepackage{bm}
\usepackage{bbm}
\usepackage{algorithmic}

\usepackage[accsupp]{axessibility}  

\usepackage[pagebackref,breaklinks,colorlinks,citecolor=eccvblue]{hyperref}

\usepackage{orcidlink}

\begin{document}

\title{Multi-Memory Matching for Unsupervised Visible-Infrared Person Re-Identification} 


\author{Jiangming Shi\inst{1}\orcidlink{0000-0003-3817-0497} \and
Xiangbo Yin\inst{2}\orcidlink{0000-0002-8599-909X}\and
Yeyun Chen\inst{1}\orcidlink{0000-0001-7234-0993}\and
Yachao Zhang\inst{3}\orcidlink{0000-0002-6153-5004}\and\\
Zhizhong Zhang\inst{4,5}\orcidlink{0000-0001-6905-4478}\and
Yuan Xie\inst{4}$^\star$\orcidlink{0000-0001-6945-7437}\and
Yanyun Qu\inst{1,2}\thanks{Corresponding author}\orcidlink{0000-0002-8926-4162} }

\authorrunning{Jiangming Shi et al.}

\institute{Institute of Artificial Intelligence, Xiamen University, Xiamen, China \\
\and
School of Informatics, Xiamen University, Xiamen, China\and
Tsinghua Shenzhen International Graduate School, Tsinghua University, Shenzhen, China\and
 School of Computer Science and Technology, East China Normal University,
Shanghai, China\\ \and
Shanghai Key Laboratory of Computer Software Evaluating and Testing, Shanghai, China
\email{jiangming.shi@outlook.com}, \email{yyqu@xmu.edu.cn}\\
\email{\{yxie,zzzhang\}@cs.ecnu.edu.cn}\\
\url{https://github.com/shijiangming1/MMM}}
\maketitle

\begin{abstract}
Unsupervised visible-infrared person re-identification (USL-VI-ReID) is a promising yet highly challenging retrieval task. The key challenges in USL-VI-ReID are to accurately generate pseudo-labels and establish pseudo-label correspondences across modalities without relying on any prior annotations. Recently, clustered pseudo-label methods have gained more attention in USL-VI-ReID. However, most existing methods don't fully exploit the intra-class nuances, as they simply utilize a single memory that represents an identity to establish cross-modality correspondences, resulting in noisy cross-modality correspondences. To address the problem, we propose a Multi-Memory Matching (MMM) framework for USL-VI-ReID. We first design a simple yet effective Cross-Modality Clustering (CMC) module to generate the pseudo-labels through clustering together both two modality samples. To associate cross-modality clustered pseudo-labels, we design a Multi-Memory Learning and Matching (MMLM) module, ensuring that optimization explicitly focuses on the nuances of individual perspectives and establishes reliable cross-modality correspondences. Finally, we design a Soft Cluster-level Alignment (SCA) loss to narrow the modality gap while mitigating the effect of noisy pseudo-labels through a soft many-to-many alignment strategy. Extensive experiments on the public SYSU-MM01 and RegDB datasets demonstrate the reliability of the established cross-modality correspondences and the effectiveness of MMM.

\end{abstract}

\section{Introduction}
\label{sec:intro}

Person re-identification (ReID) is a retrieval task, which aims to match the same person across different cameras, serving critical roles in video surveillance applications like intelligent security~\cite{SSG, MMT} and human analysis~\cite{CADRL, TQFAT}. However, in low-light conditions, the images captured by visible cameras are far from satisfactory, which renders methods~\cite{SGPT, DPM, LAW, CCL} that primarily focus on matching visible images less effective. Fortunately, smart surveillance cameras that can switch from visible to infrared modes in poor lighting environments have become widespread, driving the development of visible-infrared person re-identification (VI-ReID) for the 24-hour surveillance system.

\begin{figure}[tb]
    \centering	\includegraphics[width=0.7\linewidth]{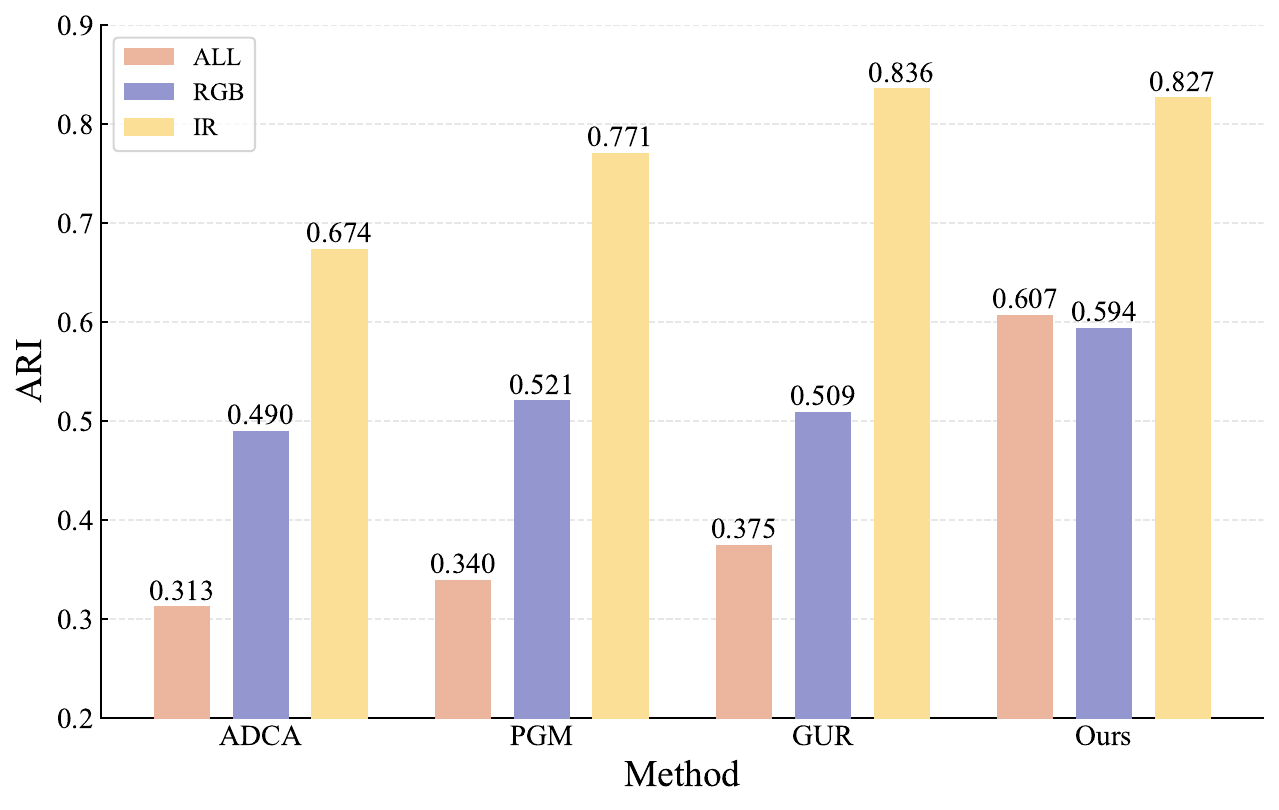}
    \caption{Comparision with different methods on ARI. The ARI indicates the Adjusted Rand Index, which is a similarity measure between two clusterings. The ALL category represents the ARI values of overall pseudo-labels, composed of visible and infrared pseudo-labels, and serves as a metric for evaluating the reliability of cross-modality correspondences.}
	\label{Fig:Intro}
\end{figure}

VI-ReID aims at retrieving infrared images of the same person when provided with a visible person image, and vice versa~\cite{TOP-ReID,DART,CAJD}. Many VI-ReID methods~\cite{MT, FMCNet, CNL,OII} have shown promising progress. However, these methods are based on well-annotated cross-modality data, which is time-consuming and labor-intensive, thereby limiting the practical application of supervised VI-ReID methods in real-world scenarios.

To free the toilsome label process and speed the automation of VI-ReID, several unsupervised VI-ReID (USL-VI-ReID) methods~\cite{RPL,SDCL,he2023efficient,yang2023towards} have been proposed, which try to establish cross-modality correspondences by clustering pseudo-labels and have achieved fairly good performance. However, the reliability of pseudo-labels and cross-modality correspondences in USL-VI-ReID still is untouched. We argue the problem is critical to the credibility of USL-VI-ReID. To measure the reliability, we introduce the Adjusted Rand Index (ARI) metric~\cite{ARI}, which is a widely recognized metric for clustering evaluation. The larger the ARI value, the better it reflects the degree of overlap between the clustered results and the ground-truth labels. More detailed explanations are presented in \textbf{supplementary materials}. In Fig.~\ref{Fig:Intro}, RGB and IR categories denote the ARI values of visible and infrared pseudo-labels, which can measure the quality of visible and infrared pseudo-labels. Interestingly, we unveil a paradoxical phenomenon: the reliability of cross-modality correspondences in previous methods stands questioned, notwithstanding their demonstrated efficacy, as depicted in Fig.~\ref{Fig:Intro} and Tab.~\ref{tab:us}. This conundrum may arise from the reality that individuals, despite bearing unique identities, manifest overlapping attributes, which tend to merge more closely due to noisy correspondences. Although this amalgamation of similar characteristics can inadvertently heighten the similarity across cross-modality features, which may lead to increased significant challenges in the precise retrieval of specific persons from a densely populated gallery.

To reduce the noisy cross-modality correspondences in USL-VI-ReID, we develop a novel Multi-Memory Matching~(MMM) framework. Multi-memory can store a wider array of distinct characteristics for an identity. For example, Memory 1 can retain front-facing attributes, Memory 2 can capture rear-facing attributes. In short, multi-memory supports a more diverse representation, which is beneficial for the establishment of cross-modality correspondences.
Specifically, we propose a Cross-Modality Clustering (CMC) module to generate pseudo-labels. Unlike previous methods, we not only cluster intra-modality samples but also cluster inter-modality samples to learn modality-invariant features. We note that the existing methods typically rely on a single memory to represent individual characteristics and establish cross-modality correspondences. However, a single memory may not capture all individual nuances, including perspective, attire, and other factors,  which naturally leads to poor cross-modality correspondences. Therefore, we design a Multi-Memory Learning and Matching (MMLM) module to obtain reliable cross-modality correspondences. We subdivide single memory into multi-memory for a single identity by sub-cluster and compute a cost matrix for multi-memory. To reduce the discrepancy between the two modalities, we propose the Soft Cluster-level Alignment (SCA) loss to narrow the modality gap through soft cluster-level intra- and inter-modality alignment. MMM can achieve fairly good quality of pseudo-labels and cross-modality correspondences compared with several USL-VI-ReID methods, as shown in Fig.~\ref{Fig:Intro}.

The main contributions are summarized as follows:
\begin{itemize}

    \item We introduce the ARI metric to evaluate the quality of pseudo-labels and cross-modality correspondences. We observe a curious phenomenon: the cross-modality correspondences of previous methods are not reliable, though they achieve good performance.
    
    \item We design a novel Multi-Memory Matching (MMM) framework for unsupervised VI-ReID, which exploits the individual nuances to effectively establish reliable cross-modality correspondences.
    
    \item We introduce two effective modules and one loss: Cross-Modality Clustering (CMC), Multi-Memory Learning and Matching (MMLM), and Soft Cluster-level Alignment (SCA). They facilitate the generation of pseudo-labels, establish reliable cross-modality correspondences, and narrow the discrepancy between two modalities while mitigating the influence of noisy pseudo-labels.
\end{itemize}

\section{Related Work}
\label{sec:related}
\subsection{Supervised Visible-Infrared Person ReID}
Visible-infrared person ReID is a challenging cross-modality image retrieval problem~\cite{I2W,ABS}. Many works have been proposed to alleviate the large cross-modality gap for VI-ReID, which can be broadly categorized into two classes: image-level alignment and feature-level alignment. The image-level alignment methods~\cite{ CMPG, TOP-ReID} try to generate cross-modal images to excavate modality-invariant information. Moreover, several methods~\cite{VIS, MMN, SMCL} introduce an auxiliary modality to assist the cross-modality retrieval task.
The feature-level alignment methods~\cite{ CAJ, DART, PSA, DCLNet,UFineBench} mainly map cross-modal features into a shared feature space to reduce cross-modal differences. For example, SGIEL~\cite{SGIEL} separates shape-related features from shape-erasure features through orthogonal decomposition to improve the diversity and identification of the learned representations for VI-ReID.
However, the above methods heavily rely on large-scale cross-modality data annotation, which is quite expensive and time-consuming.

\subsection{Unsupervised Single-Modality Person ReID}
Existing unsupervised single-modality person ReID (USL-ReID) methods can be roughly categorized into domain translation-based methods and clustering-based methods. The domain translation-based methods \cite{SSG, MMT, SPCL, MEB, CIFL} try to transfer the knowledge from the labeled source domain to the unlabeled target domain for USL-ReID. Compared with the former, the clustering-based methods \cite{SSL, MLC, CCL, ICE} are more challenging, which are trained directly on the unlabeled target domain. The common idea of clustering-based methods is using clustering algorithms~\cite{DBSCAN} to generate pseudo-labels to train a ReID model. Pseudo-labels inevitably contain noise, so it is challenging to assign the correct label to each unlabeled image. Recently, Cluster-Contrast \cite{cluster-contrast} performs contrastive learning at the cluster level with a uni-centroid. However, a uni-centroid cannot represent a cluster well. Therefore, MCRN~\cite{MCRN} and DCMIP~\cite{DCMIP} store multi-centroid representations to completely represent a cluster. Their multi-centroids are obtained through initialization, but these divisions do not accurately represent the real distribution.
Although the above methods perform well on USL-ReID, they are not suitable for solving the USL-VI-ReID due to the large cross-modality gap.

\subsection{Unsupervised Visible-Infrared Person ReID}
The challenge of unsupervised VI-ReID (USL-VI-ReID) is establishing reliable cross-modality correspondence. 
H2H~\cite{H2H} and OTLA~\cite{OTLA} use a well-annotated labeled source domain for pre-training to solve the USL-VI-ReID. Inspired by Cluster-Contrast \cite{cluster-contrast} for USL-ReID,  some clustering-based methods~\cite{PPLR, cheng2023unsupervised, he2023efficient, PGMAL, yang2023towards} are proposed for USL-VI-ReID, they try to establish cross-modality correspondence by clustering pseudo-labels. Recently, it has been shown that the Large-scale Vision-Language Pre-training model, naturally excels in producing textual descriptions for images. To this end, CCLNet~\cite{CCLNet} leverages the text information from CLIP to improve the USL-VI-ReID task. However, none of the above methods evaluate the reliability of cross-modality correspondence,  indeed, their cross-modality correspondence is not reliable. Our method aims to investigate how to establish more reliable cross-modality correspondence for USL-VI-ReID.

\begin{figure*}[tb]
    \centering	\includegraphics[width=1.0\linewidth]{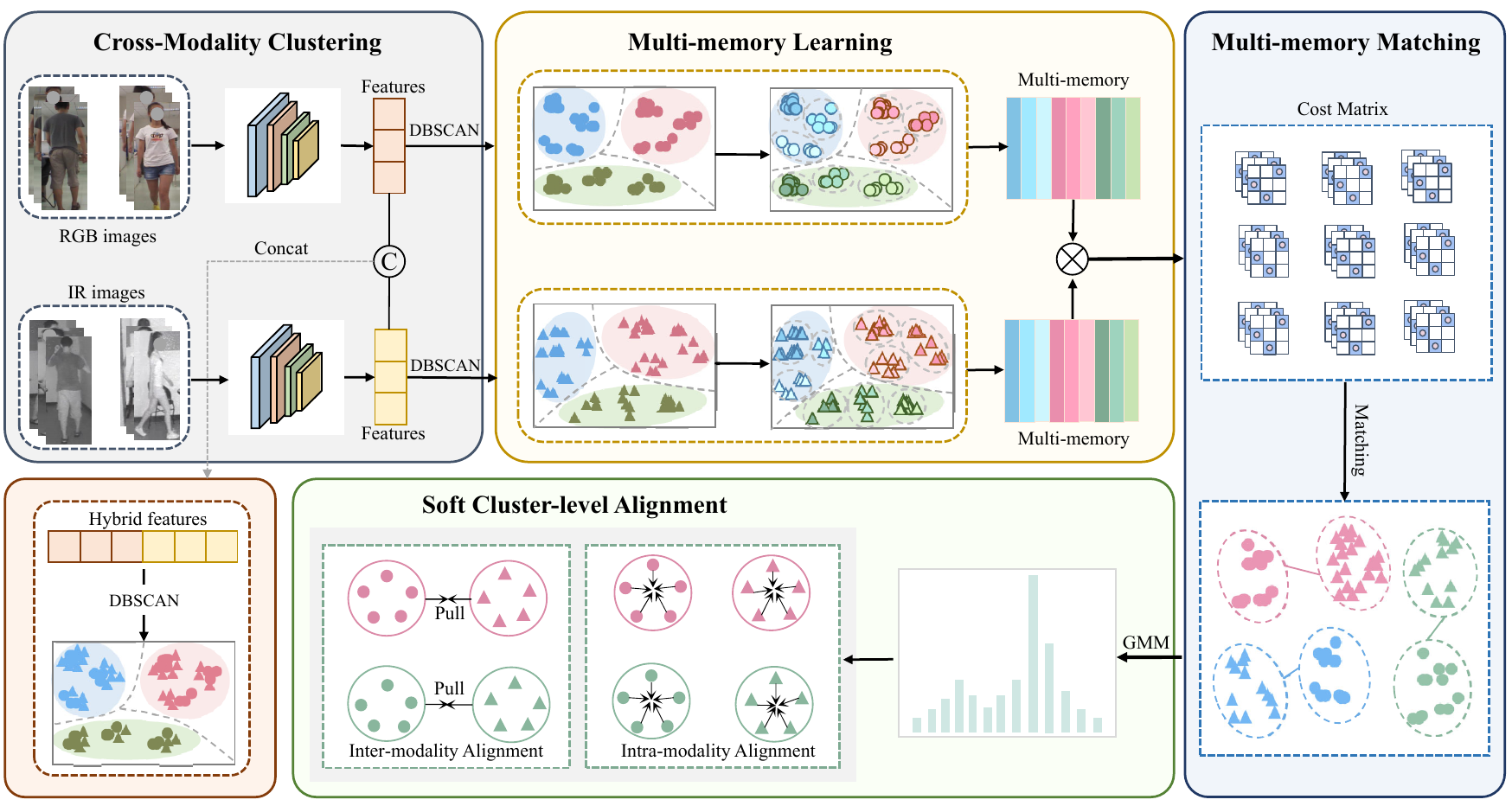}
    \caption{The pipeline of MMM. Different colors indicate different persons, $\bigcirc$ and $\bigtriangleup$ indicate visible and infrared features. It contains the Cross-Modality Clustering module (Baseline, described in Sec.~\ref{CMC}) and two key novel components: Multi-Memory Learning and Matching (MMLM, described in Sec.~\ref{MMLM}) and Soft Cluster-level Alignment (SCA, described in Sec.~\ref{SCA}). 
}
	\label{Fig:Framework}
\end{figure*}

\section{Methodology}
\label{sec:method}
The framework of MMM is illustrated in Fig.~\ref{Fig:Framework}. We begin by employing the Cross-Modality Clustering (CMC) module to generate pseudo-labels. Building upon CMC, we propose a novel Multi-Memory Learning and Matching (MMLM) module to effectively establish cross-modality correspondences. Finally, we propose the Soft Cluster-level Alignment (SCA) loss to narrow the gap between two modalities while mitigating the impact of noisy pseudo-labels through soft cluster-level intra- and inter-modality alignment.

\subsection{Notation Definition}
{\label{setting}}
Suppose we have a USL-VI-ReID dataset denoted as $D=\{V,R\}$. Here, $V = \{v_i\}^{N}_{i=1}$ represents the visible images with $N$ samples, and $R = \{r_i\}^{M}_{i=1}$ denotes the infrared images with $M$ samples. We initialize their pseudo-labels as $Y^t$, where $t \in \{v,r\}$. Let $N_p$ and $M_p$ represent the number of visible and infrared samples with ID $p$, where $p \in \{1,2,...,P^t\}$ and $P^t$ is the total number of person identities for modality $t$. The respective feature sets of these images are denoted as \(F^v = \{f^v_1, f^v_2, \ldots, f^v_{N}\}\) for visible samples and \(F^r = \{f^r_1, f^r_2, \ldots, f^r_{M}\}\) for infrared samples, respectively. 
Our goal is to develop a cross-modality person ReID model without utilizing any labels.

\subsection{Cross-Modality Clustering  }
\label{CMC}
Most USL-VI-ReID methods typically use clustering algorithms to generate pseudo-labels. Following this paradigm, we employ the DBSCAN algorithm~\cite{DBSCAN} to generate pseudo-labels for all images, as described:
\begin{equation}
{{Y}^{t}}=DBSCAN(F^t).
\end{equation}

Unlike previous methods, we not only cluster intra-modality samples $(t=v~or~t=r)$ but also cluster inter-modality samples $(t=\{v,r\})$ to indirectly build cross-modality correspondence. 

At the beginning of every training iteration, we calculate and store the memory for each cluster as follows:
\begin{equation}
   \label{Q1}
   \boldsymbol{C}_{V^{p}} = \frac{1}{N_{p}} \sum_{i=1}^{N_{p}} {f}(V_{i}^p),
\end{equation}
\begin{equation}
   \label{Q2}
   \boldsymbol{C}_{R^{p}} = \frac{1}{M_{p}} \sum_{i=1}^{M_{p}} {f}(R_{i}^p),
\end{equation}
\begin{equation}
   \label{Q3}
   \boldsymbol{C}_{{VR}^{p}} = \frac{1}{A_{p}} \sum_{i=1}^{A_{p}} {f}(VR_{i}^p),
\end{equation}
where $f(\cdot)$ is a function designated for extracting features from images across diverse modalities. We use superscripts to denote specified identity, $V^{p}$ and $R^{p}$ denote the visible and infrared modality of the same identity sample sets with ID $p$, respectively. ${VR}^{p}$ represents the combined set of both modalities with $A_{p}$ samples of the same ID $p$. 

Then, we optimize the feature extractor using ClusterNCE~\cite{cluster-contrast} loss, computed as:
 \begin{equation}
L_{V}=-\log\frac{\exp\left({C_{V}^{+}}\cdot F^v/\tau\right)}{\sum_{p=1}^{P^{v}}\exp\left({C_{V^p}}\cdot F^v/\tau\right)},
 \end{equation}
  \begin{equation}
L_{R}=-\log\frac{\exp\left({C_{R}^{+}}\cdot F^r/\tau\right)}{\sum_{p=1}^{P^{r}}\exp\left({C_{R^p}}\cdot F^r/\tau\right)},
 \end{equation}
  \begin{equation}
L_{VR}=-\log\frac{\exp\left({C_{{VR}}^{+}}\cdot [F^v, F^r]/\tau\right)}{\sum_{p=1}^{P^{v,r}}\exp\left({C_{{VR}^p}}\cdot [F^v, F^r]/\tau\right)},
 \end{equation}
 where $C^{+}$ is the positive memory representation and the $\tau$ is a temperature hyper-parameter.
 
The CMC loss is defined as:
 \begin{equation}
L_{CMC}=L_{{V}}+L_{{R}}+L_{{VR}}.
 \end{equation}

\subsection{Multi-Memory Learning and Matching}
\label{MMLM}
The CMC optimizes the feature extractor using a single memory,  but a single memory may not fully capture individual nuances, such as perspective and attire. Moreover, the CMC does not directly establish relations between the two modalities, thereby limiting its effectiveness in cases with significant modality discrepancies. To more effectively capture individual nuances and bridge the gap between the visible and infrared modalities, we propose the Multi-Memory Learning and Matching (MMLM) module, which mines a holistic representation and establishes reliable cross-modality correspondences. Specifically, we further subdivide single memory into multi-memory for a single identity, which can be formulated as a sub-cluster:

\begin{equation}
\underset{{F_{C_{V^p_i}}}}{\min}\{ \sum_{i=1}^{n} \{\left\|{f^v}-K_{C_{V^p_i}}\right\|_{2}^{2},\forall f^v \in F_{C_{V^p_i}}\}\},
\end{equation}
\begin{equation}
\underset{{F_{C_{R^p_i}}}}{\min}\{ \sum_{i=1}^{n} \{\left\|{f^r}-K_{C_{R^p_i}}\right\|_{2}^{2},\forall f^r \in F_{C_{R^p_i}}\}\},
\end{equation}

where $F_{C_{V^p_i}}$ and $F_{C_{R^p_i}}$ represent the $i$-th visible and infrared feature sets of ID $p$, respectively. $n$ is the number of memories for a single identity.
\begin{equation}
K_{C_{V^p_i}}=\frac{1}{|F_{C_{V^p_i}}|} \sum_{f^v \in F_{C_{V^p_i}}} f^v,
\end{equation}
\begin{equation}
K_{C_{R^p_i}}=\frac{1}{|F_{C_{R^p_i}}|} \sum_{f^r \in F_{C_{R^p_i}}} f^r,
\end{equation}
where $K_{C_{V^p}}$ and $K_{C_{R^p}}$ represent the visible and infrared multi-memory of ID $p$. 
 
By employing the multi-memory learning strategy, we achieve more diverse memories for a single identity. However, these memories still exhibit a strong implicit correlation with the modality, which negatively impacts the establishment of cross-modality correspondences. Inspired by PGM~\cite{PGMAL}, we transform the cross-modality multi-memory matching problem into a weighted bipartite graph matching. The goal is to match each visible cluster with the corresponding identity infrared cluster while minimizing the cost, which is formulated as follows:

 \begin{equation}
\begin{array}{c}
\underset{{Q}}{\min } M^{T} {Q}  \\
\text { s.t. } \forall p \in [P^v], \forall p' \in [P^r] : Q_{p}^{p'} \in\{0,1\}, \\
\forall p \in [P^v]: \underset{{p' \in [P^r]}}\sum Q_{p}^{p'} \leq 1, \\
\forall p' \in [P^r] : \underset{{p \in [P^v}]} \sum Q_{p}^{p'}=1,
\end{array}
 \end{equation}

 where ${Q}=\left\{Q_{p}^{p'}\right\} \in \mathbb{R}^{P^v \times P^r \times 1} $ indicates whether $K_{V^p}$ and $K_{R^{p'}}$  belong to the same person $\left(Q_{p}^{p'}=1\right) $ or not  $\left(Q_{p}^{p'}=0\right)$. $M$ and $[P^t]$ denote cost matrix and $\{1,\dots,P^t\}$, respectively. We design a simple yet effective cost expression for cross-modality multi-memory matching as follows:

 \begin{equation}
M(K_{C_{V^p}}, K_{C_{R^{p'}}}) = \sum_{i=1}^{n} \min_{j \in \{1, \cdots, n\}} \|K_{V^p_i}, K_{R^{p'}_j}\|_2,
 \end{equation}
Finally, we transfer the infrared pseudo-labels to the visible pseudo-labels, and the visible pseudo-labels are updated by:
\begin{equation}
{Y^v}:=QY^r.
\end{equation}

\subsection{Soft Cluster-level Alignment}
\label{SCA}
Pseudo-labels inherently contain noise, a problem that is not exempt even in human annotations~\cite{DART}, leading to a reduction in performance. The method~\cite{ACL} illustrated that deep neural networks initially learn from simple samples before accommodating noisy labels. Building on this insight, we assess the confidence associated with each label. To do so, we employ a two-component Gaussian Mixture Model (GMM) to model the loss distribution:

\begin{equation}
{L}_{ID}^v= {-\log p\left( {{Y}^v}  \mid C\left(F^v\right)\right)},
\label{idloss}
\end{equation}
\begin{equation}
p({L}^v_{ID}\mid \theta)=\sum_{k=1}^{2}\pi_{k}\phi({L}_{ID}^v \mid k),
\end{equation}
where $C(\cdot)$ acts as an identity classifier. $\pi_{k}$ represents the mixture coefficient, while $\phi({L}^{v}_{ID}\mid k)$ denotes the probability density of the $k$-th component.

Subsequently, the confidence is determined by computing its posterior probability, detailed as:
\begin{equation}
W^v=p\left(k \mid {L}^v_{ID}\right),
\label{pro_w}
\end{equation}
where $k$ refers to the Gaussian component with a smaller mean, while $p\left(k \mid {L}^v_{ID}\right)$ indicates the responsiveness of ${L}^v_{ID}$ to the $k$-th component.
In the same way, we can obtain the confidence $W^r$ and $W^{vr}$.

 To penalize the noise during optimization, the memories in  Eq.~(\ref{Q1}), (\ref{Q2}), (\ref{Q3}) are updated by:
\begin{equation}
   {C}_{V^{p}} := \frac{1}{N_{p}} \sum_{i=1}^{N_{p}} {f}(V_{i}^p)W_{V^p_i},
   \label{centerOfflineRGB1}
\end{equation}
\begin{equation}
   {C}_{R^{p}} := \frac{1}{M_{p}} \sum_{i=1}^{M_{p}} {f}(R_{i}^p)W_{R^p_i},
   \label{centerOfflineIR2}
\end{equation}
\begin{equation}
   {C}_{{VR}^{p}} := \frac{1}{A_{p}} \sum_{i=1}^{A_{p}} {f}(VR_{i}^p)W_{VR^p_i},
   \label{centerOfflineVR1}
\end{equation}
where $W_{V^p_i}$, $W_{R^p_i}$, and $W_{VR^p_i}$ denote the confidences of samples $V^p_i$, $R^p_i$, and $VR^p_i$, respectively.

To reduce the intra-modality discrepancy, we employ the distilled ${C}_{V^{p}}$ and ${C}_{R^{p}}$ to align every sample of ID $p$ to its corresponding memory in each modality. The cluster-level intra-modality alignment loss ${L}_{Intra}$ is proposed as:
\begin{equation}
\begin{split}
   {L}_{Intra} &= {L}_{Intra}^{V}+ {L}_{Intra}^{R} \\
   &= \sum_{p=1}^{P^v} \sum_{f^v \in F_{p}^v} \left \|   f^v -  {C}_{V^{p}} \right \|_{2}^{2}\\
   &+ \sum_{p=1}^{P^r} \sum_{f^r \in F_{p}^r} \left \|   f^r -  {C}_{R^{p}}\right \|_{2}^{2},
\end{split}
\label{LMI}
\end{equation}
where ${F}^v_{p}$, ${F}^r_{p}$ denote visible feature and infrared feature sets of ID $p$, respectively.

Since VI-ReID is a many-to-many matching problem, we propose cluster-level inter-modality alignment loss, which forces the feature distribution of the  samples from the visible modality to be similar to the feature distribution of the  samples from the infrared modality and vice versa by:
  \begin{equation}
 \begin{split}
   {L}_{Inter}&= {L}_{Inter}^{V}+ {L}_{Inter}^{R}\\
   &= \frac{1}{P} \sum_{p=1}^{P} (\frac{1}{2}  D(F^v_p, sg(F^r_p) )\\
   &\quad\quad\quad + \frac{1}{2} D( F^r_p, sg(F^v_p)) ),
   \label{LMU}
\end{split}
 \end{equation}
where $sg(\cdot)$ represents the stop-gradient operation, and $D(i, j)$ represents the distance between distributions $i$ and $j$. $P$ is $\min(P^v,P^r)$. In this paper, we employ the squared Maximum Mean Discrepancy (MMD$^{2}$) \cite{MMD} to quantify the discrepancy between distributions. MMD$^{2}$ is a commonly used non-parametric metric in domain adaptation and has been observed to outperform other metrics, such as KL divergence in empirical studies, MMD$^{2}$ is constructed as:
\begin{equation}
 \begin{split}
   \text{MMD}^{2}({F}^r_{p}, {F}^v_{p}) 
  &= \frac{1}{|{F}^r_{p}|^2} \sum_{{f}^r_i \in{F}^r_{p}} \sum_{{f}^r_j\in{F}^r_{p}} z({f}^r_i, {f}^r_j)\\
   &+ \frac{1}{|{F}^v_{p}|^2} \sum_{{f}^v_i \in{F}^v_{p}}\sum_{{f}^v_i \in{F}^v_{p}} z({f}^v_i, {f}^v_j)\\
  &- \frac{2}{|{F}^r_{p}||{F}^v_{p}|} \sum_{{f}^r_i \in{F}^r_{p}}\sum_{{f}^v_j \in{F}^r_{p}} z({f}^r_i, {f}^v_j),
 \end{split}
\end{equation}
 where $z( s,  s') = \exp ( \frac{- \left \| \boldsymbol s- \boldsymbol s'\right \|_{2}^{2}}{2\sigma^{2}} )$ is a Gaussian kernel. 

 The SCA loss is defined as:
 \begin{equation}
L_{SCA}=\lambda_{Intra}L_{{Intra}}+\lambda_{Inter}L_{{Inter}},
 \end{equation}
 where $\lambda_{Intra}$ and $\lambda_{Inter}$ are the balancing weights. 

 {\noindent \bfseries  Overall Loss.}~The total loss for training the model is defined by the following equation: 
\begin{equation}
L_{overall}=L_{CMC}+L_{SCA}.
\end{equation}

\begin{table*}[!h] \small
        \caption{Comparisons with state-of-the-art methods on SYSU-MM01 and RegDB, \ie, supervised visible-infrared person ReID (SVI-ReID), semi-supervised visible-infrared person ReID (SSVI-ReID) and unsupervised visible-infrared person ReID (USL-VI-ReID). All methods are measured by Rank-1 (\%) and mAP (\%). GUR* denotes the results without camera information.}
	\label{tab:us}
	\centering
	\resizebox{\textwidth}{!}{
		\begin{tabular}{c|c|c|c|c|c|c|c|c|c|c}
			\hline
                \multicolumn{3}{c|}{\multirow{2}{*}{Settings}} & \multicolumn{4}{c|}{SYSU-MM01} & \multicolumn{4}{c}{RegDB} \\ \cline{4-11}
                \multicolumn{3}{c|}{} & \multicolumn{2}{c|}{All Search} & \multicolumn{2}{c|}{Indoor Search} & \multicolumn{2}{c|}{Visible2Thermal} & \multicolumn{2}{c}{Thermal2Visible}\\
			\hline
                Type & Method & Venue & Rank-1 & mAP & Rank-1 & mAP & Rank-1
 & mAP & Rank-1 & mAP\\
                \hline
                
               \multirow{7}{*}{SVI-ReID}         
                ~ & AGW~\cite{AGW} & TRAMI'\textcolor{blue}{21} & 47.5 & 47.7 & 54.2 & 63.0 & 70.1 & 66.4 & 70.5 & 65.9 \\  
                ~ & NFS~\cite{NFS} & CVPR'\textcolor{blue}{21} & 56.9 & 55.5 & 62.8 & 69.8 & 80.5 & 72.1 & 78.0 & 69.8 \\     
                ~ & LbA~\cite{LbA} & ICCV'\textcolor{blue}{21} & 55.4 & 54.1 & 58.5 & 66.3 & 74.2 & 67.6 & 72.4 & 65.5 \\ 
                ~ & CAJ~\cite{CAJ} & ICCV'\textcolor{blue}{21} & 69.9 & 66.9 & 76.3 & 80.4 & 85.0 & 79.1 & 84.8 & 77.8 \\  
                ~ & DART~\cite{DART} & CVPR'\textcolor{blue}{22} & 68.7 & 66.3 & 72.5 & 78.2 & 83.6 & 75.7 & 82.0 & 73.8 \\
                ~ & DEEN~\cite{DEEN} & CVPR'\textcolor{blue}{23} & 74.7 & 71.8 & 80.3 & 83.3 & 91.1 & 85.1 & 89.5 & 83.4 \\
                ~ & PartMix~\cite{PartMix} & CVPR'\textcolor{blue}{23} & 77.8 & 74.6 & 81.5 & 84.4 & 85.7 & 82.3 & 84.9 & 82.5\\
                \hline
               \multirow{3}{*}{SSVI-ReID}
               ~ & OTLA~\cite{OTLA} & ECCV'\textcolor{blue}{22} & 48.2 & 43.9 & 47.4 & 56.8 & 49.9 & 41.8 & 49.6 & 42.8\\
               ~ & TAA~\cite{taa} & TIP'\textcolor{blue}{23} & 48.8 & 42.3 & 50.1 & 56.0 & 62.2 & 56.0 & 63.8 & 56.5 \\
               ~ & DPIS~\cite{DPIS} & ICCV'\textcolor{blue}{23} & 58.4 & 55.6 & 63.0 & 70.0 & 62.3 & 53.2 & 61.5 & 52.7\\
               \hline
               \multirow{11}{*}{USL-VI-ReID}
               ~ & OTLA~\cite{OTLA} & ECCV'\textcolor{blue}{22} & 29.9 & 27.1 & 29.8 & 38.8 & 32.9 & 29.7 & 32.1 & 28.6\\ 
               ~ & ADCA~\cite{ADCA} & MM'\textcolor{blue}{22} & 45.5 & 42.7 & 50.6 & 59.1 & 67.2 & 64.1 & 68.5 & 63.8\\
               ~ & {ADCA+MMM} & - & {49.7} & {44.7} & {56.2} & {62.5} & {77.8} & {70.9} & {76.5} & {69.1} \\           
               ~ & NGLR~\cite{cheng2023unsupervised} & MM'\textcolor{blue}{23} & 50.4 & 47.4 & 53.5 & 61.7 & 85.6 & 76.7 & 82.9 & 75.0\\
               ~ & MBCCM~\cite{he2023efficient} & MM'\textcolor{blue}{23} & 53.1 & 48.2 & 55.2 & 62.0 & 83.8 & 77.9 & 82.8 & 76.7\\     
               ~ & CCLNet~\cite{CCLNet} & MM'\textcolor{blue}{23} & 54.0 & 50.2 & 56.7 & 65.1 & 69.9 & 65.5 & 70.2 & 66.7\\
               ~ & PGM~\cite{PGMAL} & CVPR'\textcolor{blue}{23} & 57.3 & 51.8 & 56.2 & 62.7 & 69.5 & 65.4 & 69.9 & 65.2\\
             ~ & CHCR~\cite{CHCR} & TCSVT'\textcolor{blue}{23} & 59.5 & 59.1 & - & - & 69.3 & 64.7 & 70.0 & 65.9\\     
              ~ & GUR*~\cite{yang2023towards} & ICCV'\textcolor{blue}{23} & 61.0 & 57.0 & 64.2 & 69.5 & 73.9 & 70.2 & 75.0 & 69.9\\
             ~ & {PCLHD~\cite{PCLMP}} & {arXiv'\textcolor{blue}{24}} & {64.4} & {58.7} & {69.5} & {74.4} & {84.3} & {80.7} & {82.7} & {78.4} \\           
               \hline
             ~ & {MMM} & - & {61.6} & {57.9} & {64.4} & {70.4} & \textbf{89.7} & {80.5} & {85.8} & {77.0} \\ 
              ~ & \textbf{MMM+PCLHD} & \textbf{-} & \textbf{65.9} & \textbf{61.8} & \textbf{70.3} & \textbf{74.9} & {89.6} & \textbf{83.7} & \textbf{87.0} & \textbf{80.9} \\ 
               \hline 
		\end{tabular}
	}
\end{table*}

\section{Experiments}
\label{sec:experiment}
In this section, we conduct comprehensive experiments to verify the effectiveness of MMM. First, we compare MMM with several state-of-the-art methods under three settings, \ie, supervised visible-infrared person ReID (SVI-ReID), semi-supervised visible-infrared person ReID (SSVI-ReID) and unsupervised visible-infrared person ReID (USL-VI-ReID). After that, we perform ablation studies to evaluate the effectiveness of each module in MMM. Finally, we perform a discussion and analysis of the hyper-parameters and visualization. If not specified, we conduct analysis experiments on SYSU-MM01 in the single-shot \& all-search mode.

\subsection{Experimental Setting}{\label{setting}}
{\noindent \bfseries Dataset.}~We evaluate MMM on two benchmarks, \ie, \textbf{SYSU-MM01}~\cite{sysu} and \textbf{RegDB}~\cite{regdb}. SYSU-MM01 is a large-scale visible-infrared person ReID dataset, which is collected from four visible cameras and two infrared cameras in both indoor and outdoor scenes. RegDB is a relatively small dataset, which is collected by one visible and one infrared camera in a dual-camera system.

{\noindent \bfseries Evaluation Protocols.}~Cumulative Matching Characteristics~\cite{EP} and mean Average Precision (mAP) are adopted as the evaluation metrics on two datasets to evaluate the performance of MMM quantitatively. For fair comparisons, we report the results of all-search mode and indoor-search mode with the official code on SYSU-MM01. Following~\cite{CAJ}, We also report the results on RegDB by randomly splitting the training and testing set 10 times in visible-to-thermal and thermal-to-visible modes.

\subsection{Implementation Details} 
We adopt ResNet50, which is initialized with the ImageNet pre-trained weights, as the shared backbone to extract 2048d features. MMM is implemented in PyTorch. The total number of training epochs is 80. At each training step, we randomly sample 8 IDs, of which 4 visible and 4 infrared images are chosen to formulate a batch. Training images are resized to $288\times144$ and random horizontal flipping and random crop are used for data augmentation~\cite{CAJ}. SGD optimizer is adopted to train the model with the momentum setting to 0.9 and weight decay setting to $5e-4$. The Intra module is added from the $1^{st}$ epoch and the Inter module is added from the $15^{th}$ epoch. The loss temperature $\tau$ is set to 0.05. The hyperparameters `eps' and `min\_samples' in DBSCAN are set to 0.6 and 4.

\begin{table*}[htb]
	\caption{Ablation studies on SYSU-MM01 in all search mode and indoor search mode. ``Baseline'' means the model trained only with the CMC module. Rank-R accuracy(\%) and mAP(\%) are reported.}
	\label{tab:ablation}
	\centering
	\resizebox{\textwidth}{!}{
		\begin{tabular}{c|cccc|ccccc|ccccc}
			\hline
                ~ &\multicolumn{4}{c|}{Method} & \multicolumn{5}{c|}{All Search} & \multicolumn{5}{c}{Indoor Search}\\
                \hline
               Order & Baseline & MMLM & Intra & Inter & Rank-1 & Rank-5 & Rank-10 & Rank-20 & mAP & Rank-1 & Rank-5 & Rank-10 & Rank-20 & mAP\\
                \hline
                 1 & \checkmark & ~ & ~ & ~ & 51.74 & 78.67 &  87.87 & 94.76 & 49.81 & 56.34 & 84.66 & 92.77 & 96.98 & 64.46 \\
                 2 & \checkmark & \checkmark & ~ & ~ & 55.15 & 81.65 & 90.53 & 96.46 & 52.21 & 58.76 & 85.21 & 93.06 & 97.16 & 65.47 \\
                 3 & \checkmark & \checkmark & \checkmark & ~ & 58.48 & 83.69 & 91.79 & 97.15 & 55.05 & 62.19 & 86.95 & 93.60 & 97.64 & 68.09\\
                 4 & \checkmark & \checkmark & ~ & \checkmark & 57.26 & 82.34 & 90.84 & 96.93 & 53.81 & 60.26 & 85.77 & 93.16 & 97.36 & 66.66\\
                 5 & \checkmark & \checkmark & \checkmark & \checkmark & \textbf{61.56} & \textbf{85.66} & \textbf{93.33} & \textbf{98.03} & \textbf{57.92} & \textbf{64.37} & \textbf{88.80} & \textbf{95.01} & \textbf{98.20} & \textbf{70.40}\\
                \hline                
		\end{tabular}}
\end{table*}

\subsection{Results and Analysis}
To clearly demonstrate the effectiveness of MMM, we compare MMM with several state-of-the-art methods under three settings, \ie, SVI-ReID, SSVI-ReID, and USL-VI-ReID. The quantitative results on SYSU-MM01 and RegDB are shown in Tab.~\ref{tab:us}. 

\noindent
\textbf{Comparison with SSVI-ReID Methods.}~ 
We compared MMM with three state-of-the-art SSVI-ReID methods. Notably, MMM not only outpaced these methods but did so without relying on any form of annotations. This stands in stark contrast to the SSVI-ReID methods, which rely on annotations of visible images to achieve their results.

\noindent
\textbf{Comparison with USL-VI-ReID Methods.}~Compared with eight state-of-the-art USVI-ReID methods, MMM consistently performs better than existing USL-VI-ReID methods by a significant margin. PCLHD~\cite{PCLMP} is proposed to learn more discriminative cross-modality features, and our method with PCLHD can achieve 65.9\% in Rank-1 and 61.8\% in mAP. ADCA~\cite{ADCA} with MMM also achieve consistently improved performance. Moreover, the results are surprising on RegDB, MMM improves the Rank-1 and mAP accuracy by a large margin of 15.8\% and 10.3\% compared to GUR under visible to thermal mode.

\noindent
\textbf{Comparison with SVI-ReID Methods.}~Surprisingly, MMM performs better than several supervised methods, including AGW~\cite{AGW}, NFS~\cite{NFS}, and LbA~\cite{LbA}. The results show the effectiveness of MMM. However, we have to acknowledge that there is still a certain gap between MMM and many SVI-ReID methods due to the absence of cross-modality data annotations.

The above results clearly show that MMM is effective, which highlights the significant potential of MMM in addressing USL-VI-ReID challenges.

\subsection{Ablation Study}
To further analyze the effectiveness of the Multi-Memory Learning and Matching (MMLM), the Soft Cluster-level Alignment (SCA), we conduct ablation studies on SYSU-MM01 under both all-search and indoor-search modes. The results are reported in Tab.~\ref{tab:ablation}.

\noindent
\textbf{Baseline.} Order 1 denotes that the model is trained only with the CMC module. Although it achieves a promising performance on SYSU-MM01, it does not directly establish relations between the two modalities, which limits the performance.

\noindent
\textbf{Effective of MMLM.} The effectiveness of the MMLM module is revealed by comparing Order 1 and Order 2. The MMLM improves 3.41\% in Rank-1 and 2.40\% in mAP on SYSU-MM01. The results, combined with Fig.~\ref{Fig:Intro}, demonstrate that the MMLM can help align visible and infrared pseudo-labels to establish cross-modality correspondences.

\noindent
\textbf{Effective of Intra in SCA.}
As shown in Order 3 of Tab.~\ref{tab:ablation}, the performance is improved to 58.48\% in Rank-1 and 55.05\% in mAP when adding the cluster-level intra-modality loss (Intra) in SCA, which shows the effectiveness of Intra in reducing the discrepancy of intra-modality.

\noindent
\textbf{Effective of Inter in SCA.} The cluster-level inter-modality alignment loss (Inter) is proposed to reduce the discrepancy of inter-modality, MMM can reach 57.26\% in Rank-1 and 53.81\% in mAP when adding it. Moreover, when combining Inter with Intra, MMM achieves the best performance with 61.56\% in Rank-1 and 57.92\% in mAP, which surpasses the baseline by a large margin of 9.82\% in Rank-1 and 8.11\% in mAP. 

The above results show that cluster-level intra- and inter-modality alignment loss can complement each other, which proves the effectiveness of the SCA loss.

\begin{figure*}[tb]
    \centering	\includegraphics[width=1.0\linewidth]{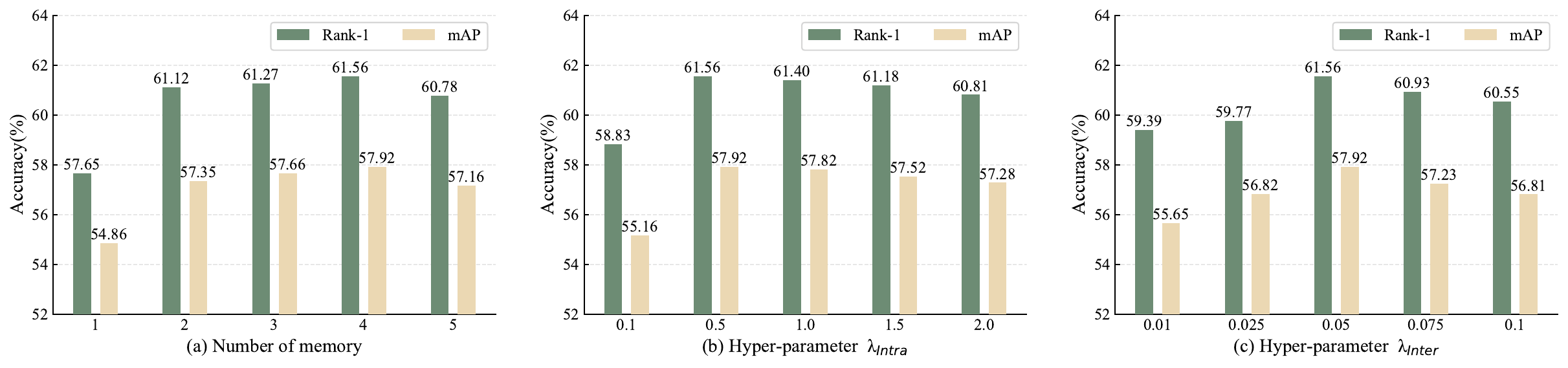}
    \caption{The effect of hyper-parameter $n$, $\lambda_{Intra}$ and $\lambda_{Inter}$ with different values on SYSU-MM01.}
	\label{Fig:hyper-parameter}
\end{figure*}

\begin{figure}[htb]
    \centering
     \includegraphics[width=0.9\linewidth]{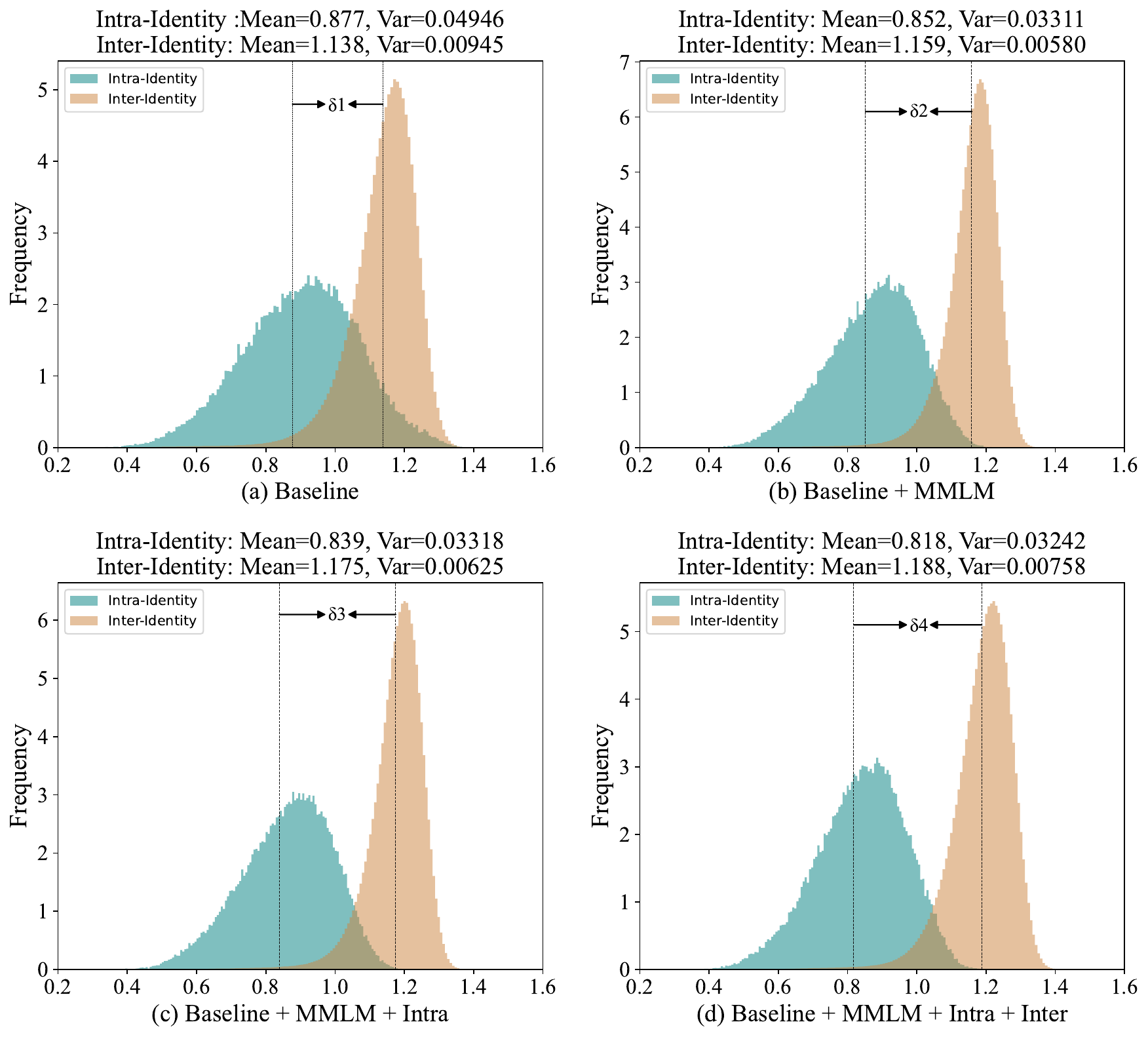}    
    \caption{The intra-identity and inter-identity distances on SYSU-MM01, where $\delta_{i}$ denotes the gap between the intra-identity distance mean and the inter-identity distance mean.}
    \label{Fig:distance}
\end{figure}

\begin{figure}[htb]
    \centering
     \includegraphics[width=1.0\linewidth]{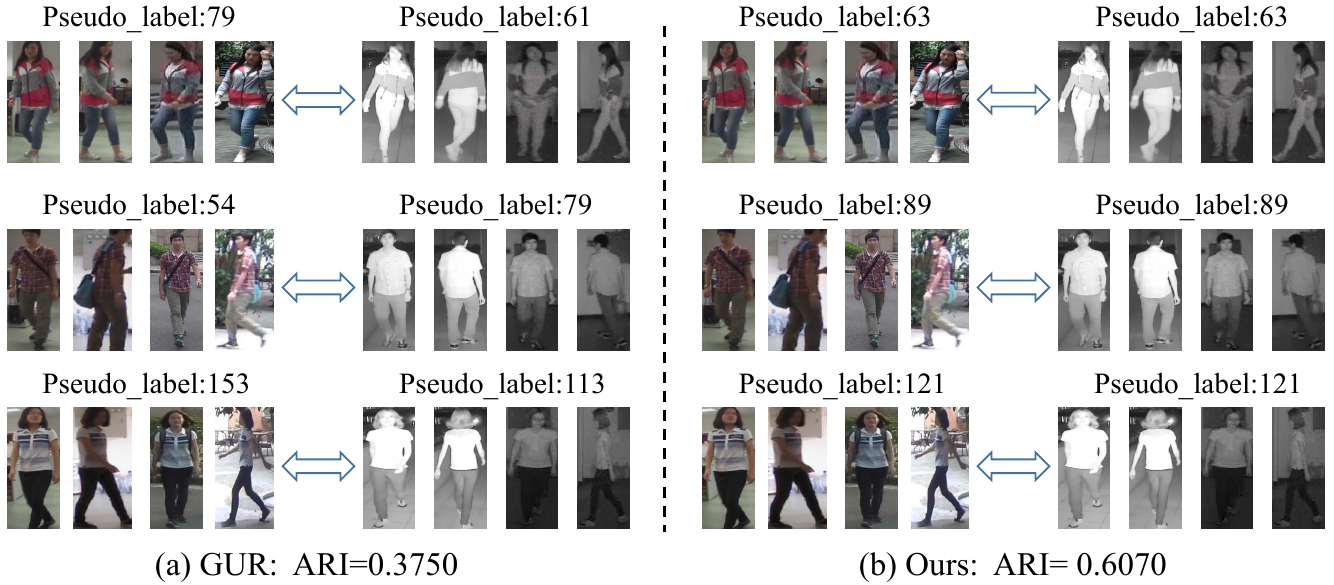}    
    \caption{The Visualization of the pseudo-labels of the same identity with different modalities. }
    \label{Fig:visualize_ARI}
\end{figure}

\subsection{Analysis of Hyper-parameters}
We analyze the key hyper-parameters of MMM on SYSU-MM01, \ie, the number of memories $n$, $\lambda_{Intra}$ and $\lambda_{Inter}$. In Fig.~\ref{Fig:hyper-parameter} (a), we vary the number of memories from 1 to 5 while keeping the $\lambda_{Intra}$ and $\lambda_{Inter}$ fixed, which shows MMM achieves the best performance with 61.56\% in Rank-1 and 57.92\% in mAP when $n=4$. Moreover, to balance the contribution between the cluster-level intra- and inter-modality alignment loss in SCA, we study the effect of $\lambda_{Intra}$ and $\lambda_{Inter}$ by fixing one and adjusting the other. To be specific, we maintain the $\lambda_{Inter}=0.05$ and tune the value of $\lambda_{Intra}$ in $[ 0.1, 0.5, 1.0, 1.5, 2.0 ]$ (Fig.~\ref{Fig:hyper-parameter} (b)), while fix the $\lambda_{Intra}=0.5$ and explore the $\lambda_{Inter}$ on different values which vary in $[ 0.01, 0.025, 0.05, 0.075, 0.1 ]$ (Fig.~\ref{Fig:hyper-parameter} (c)). We can observe that MMM achieves high accuracy under different combinations with $\lambda_{Intra}$ and $\lambda_{Inter}$, which shows the performance of MMM is not sensitive to $\lambda_{Intra}$ and  $\lambda_{Inter}$, and the best performance is achieved with $\lambda_{Intra}=0.5$ and  $\lambda_{Inter}=0.05$.

\subsection{Qualitative Analysis}
To further illustrate the effectiveness of MMM, we visualize the intra-identity and inter-identity distances on SYSU-MM01 in Fig.~\ref{Fig:distance}. As shown in Fig.~\ref{Fig:distance} (a)-(d), with the addition of the proposed methods, the means of intra-identity distances gradually decrease while the means of inter-identity distances gradually increase, which makes the intra-identity and inter-identity features distributions are pushed away ($\delta_{1} < \delta_{2} < \delta_{3} < \delta_{4}  $). The results show that MMM can effectively reduce the cross-modality distances between the same identity samples and push the distance between different identity samples far away.

Moreover, we also visualize the pseudo-labels of the same identity with different modalities, where we randomly choose 3 person identities, where each identity consists of 4 visible images and infrared images. As shown in Fig.~\ref{Fig:visualize_ARI}, 
persons of the same identity in different modalities have the same pseudo-label in MMM (right) compared with GUR (left), which shows that MMM can establish more reliable cross-modality correspondences.

\section{Conclusion}
In this paper, we introduce a metric, the Adjusted Rand Index, to measure cross-modality correspondences and clustered pseudo-labels, exploring the establishment of reliable cross-modality correspondences for USL-VI-ReID. To this end, we propose a Multi-Memory Matching (MMM) framework. Firstly, we design a Cross-Modality Clustering (CMC) module to generate pseudo-labels. Instead of previous methods, we employ multi-memory in the Multi-Memory Learning and Matching (MMLM) module to capture individual nuances and establish reliable cross-modality correspondences. Additionally, we present a Soft Cluster-level Alignment (SCA) loss to reduce the cross-modality gap while mitigating the effect of noisy pseudo-labels. Comprehensive experimental results show that MMM can establish reliable cross-modality correspondences and outperforms existing USL-VI-ReID methods on SYSU-MM01 and RegDB.

\textbf{Acknowledgments.} This work is supported by the National Natural Science Foundation of China (No. 62176224, 62222602, 62106075, 62176092, 62306165), Natural Science Foundation of Shanghai (23ZR1420400), Natural Science Foundation of Chongqing (CSTB2023NSCQ-JQX0007), China Postdoctoral Science Foundation (No. 2023M731957), CCF-Lenovo Blue Ocean Research Fund.

\clearpage  

%
%
\bibliographystyle{splncs04}
\bibliography{main}

\begin{thebibliography}{10}
\providecommand{\url}[1]{\texttt{#1}}
\providecommand{\urlprefix}{URL }
\providecommand{\doi}[1]{https://doi.org/#1}

\bibitem{ACL}
Arpit, D., Jastrzebski, S., Ballas, N., Krueger, D., Bengio, E., Kanwal, M.S., Maharaj, T., Fischer, A., Courville, A.C., Bengio, Y., Lacoste{-}Julien, S.: A closer look at memorization in deep networks. In: {ICML}. pp. 233--242 (2017)

\bibitem{ICE}
Chen, H., Lagadec, B., Br{\'{e}}mond, F.: {ICE:} inter-instance contrastive encoding for unsupervised person re-identification. In: {ICCV}. pp. 14940--14949 (2021)

\bibitem{NFS}
Chen, Y., Wan, L., Li, Z., Jing, Q., Sun, Z.: Neural feature search for rgb-infrared person re-identification. In: {CVPR}. pp. 587--597 (2021)

\bibitem{CCLNet}
Chen, Z., Zhang, Z., Tan, X., Qu, Y., Xie, Y.: Unveiling the power of clip in unsupervised visible-infrared person re-identification. In: ACM MM. pp. 3667--3675 (2023)

\bibitem{cheng2023unsupervised}
Cheng, D., Huang, X., Wang, N., He, L., Li, Z., Gao, X.: Unsupervised visible-infrared person reid by collaborative learning with neighbor-guided label refinement. ArXiv:2305.12711  (2023)

\bibitem{PPLR}
Cho, Y., Kim, W.J., Hong, S., Yoon, S.: Part-based pseudo label refinement for unsupervised person re-identification. In: {CVPR}. pp. 7298--7308 (2022)

\bibitem{cluster-contrast}
Dai, Z., Wang, G., Yuan, W., Zhu, S., Tan, P.: Cluster contrast for unsupervised person re-identification. In: ACCV. pp. 319--337 (2022)

\bibitem{DBSCAN}
Ester, M., Kriegel, H., Sander, J., Xu, X.: A density-based algorithm for discovering clusters in large spatial databases with noise. In: KDD. pp. 226--231 (1996)

\bibitem{SGIEL}
Feng, J., Wu, A., Zheng, W.: Shape-erased feature learning for visible-infrared person re-identification. In: CVPR. pp. 22752--22761 (2023)

\bibitem{SSG}
Fu, Y., Wei, Y., Wang, G., Zhou, Y., Shi, H., Huang, T.S.: Self-similarity grouping: {A} simple unsupervised cross domain adaptation approach for person re-identification. In: ICCV. pp. 6111--6120 (2019)

\bibitem{MMT}
Ge, Y., Chen, D., Li, H.: Mutual mean-teaching: Pseudo label refinery for unsupervised domain adaptation on person re-identification. In: ICLR (2020)

\bibitem{SPCL}
Ge, Y., Zhu, F., Chen, D., Zhao, R., Li, H.: Self-paced contrastive learning with hybrid memory for domain adaptive object re-id. In: NeurIPS (2020)

\bibitem{CAJD}
Gong, Y., Huang, L., Chen, L.: Person re-identification method based on color attack and joint defence. In: {IEEE/CVF} Conference on Computer Vision and Pattern Recognition Workshops. pp. 4312--4321. {IEEE} (2022)

\bibitem{PSA}
Gong, Y., Zhong, Z., Luo, Z., Qu, Y., Ji, R., Jiang, M.: Cross-modality perturbation synergy attack for person re-identification. CoRR  \textbf{abs/2401.10090} (2024)

\bibitem{MMD}
Gretton, A., Borgwardt, K.M., Rasch, M.J., Sch{\"o}lkopf, B., Smola, A.: A kernel two-sample test. The Journal of Machine Learning Research  \textbf{13}(1),  723--773 (2012)

\bibitem{he2023efficient}
He, L., Wang, N., Zhang, S., Wang, Z., Gao, X., et~al.: Efficient bilateral cross-modality cluster matching for unsupervised visible-infrared person reid. ArXiv:2305.12673  (2023)

\bibitem{ARI}
Hubert, L., Arabie, P.: Comparing partitions. Journal of classification  \textbf{2},  193--218 (1985)

\bibitem{PartMix}
Kim, M., Kim, S., Park, J., Park, S., Sohn, K.: Partmix: Regularization strategy to learn part discovery for visible-infrared person re-identification. In: CVPR. pp. 18621--18632 (2023)

\bibitem{OII}
Li, H., Ye, M., Zhang, M., Du, B.: All in one framework for multimodal re-identification in the wild. In: CVPR. pp. 17459--17469 (2024)

\bibitem{H2H}
Liang, W., Wang, G., Lai, J., Xie, X.: Homogeneous-to-heterogeneous: Unsupervised learning for rgb-infrared person re-identification. {IEEE} Trans. Image Process.  \textbf{30},  6392--6407 (2021)

\bibitem{CADRL}
Lin, L., Liu, H., Liang, J., Li, Z., Feng, J., Han, H.: Consensus-agent deep reinforcement learning for face aging. IEEE Transactions on Image Processing  (2024)

\bibitem{TQFAT}
Lin, L., Wang, T., Liu, H., Zhu, C., Chen, J.: Toward quantifiable face age transformation under attribute unbias. IEEE Transactions on Circuits and Systems for Video Technology  (2024)

\bibitem{SSL}
Lin, Y., Xie, L., Wu, Y., Yan, C., Tian, Q.: Unsupervised person re-identification via softened similarity learning. In: {CVPR}. pp. 3387--3396 (2020)

\bibitem{regdb}
Nguyen, D.T., Hong, H.G., Kim, K., Park, K.R.: Person recognition system based on a combination of body images from visible light and thermal cameras. Sensors  \textbf{17}(3), ~605 (2017)

\bibitem{CHCR}
Pang, Z., Wang, C., Zhao, L., Liu, Y., Sharma, G.: Cross-modality hierarchical clustering and refinement for unsupervised visible-infrared person re-identification. IEEE Transactions on Circuits and Systems for Video Technology pp.~1--1 (2023)

\bibitem{CIFL}
Pang, Z., Zhao, L., Liu, Q., Wang, C.: Camera invariant feature learning for unsupervised person re-identification. IEEE transactions on multimedia  \textbf{25},  6171--6182 (2022)

\bibitem{LbA}
Park, H., Lee, S., Lee, J., Ham, B.: Learning by aligning: Visible-infrared person re-identification using cross-modal correspondences. In: {ICCV}. pp. 12026--12035 (2021)

\bibitem{VIS}
Shi, J., Yin, X., Zhang, D., Qu, Y.: Visible embraces infrared: Cross-modality person re-identification with single-modality supervision. In: 2023 China Automation Congress (CAC). pp. 4781--4787. IEEE (2023)

\bibitem{PCLMP}
Shi, J., Yin, X., Zhang, Y., Zhang, Z., Xie, Y., Qu, Y.: Learning commonality, divergence and variety for unsupervised visible-infrared person re-identification. arXiv:2402.19026  (2024)

\bibitem{DPIS}
Shi, J., Zhang, Y., Yin, X., Xie, Y., Zhang, Z., Fan, J., Shi, Z., Qu, Y.: Dual pseudo-labels interactive self-training for semi-supervised visible-infrared person re-identification. In: ICCV. pp. 11218--11228 (2023)

\bibitem{DCLNet}
Sun, H., Liu, J., Zhang, Z., Wang, C., Qu, Y., Xie, Y., Ma, L.: Not all pixels are matched: Dense contrastive learning for cross-modality person re-identification. In: ACM MM. pp. 5333--5341 (2022)

\bibitem{DPM}
Tan, L., Dai, P., Ji, R., Wu, Y.: Dynamic prototype mask for occluded person re-identification. In: ACM MM. pp. 531--540 (2022)

\bibitem{SGPT}
Tan, L., Xia, J., Liu, W., Dai, P., Wu, Y., Cao, L.: Occluded person re-identification via saliency-guided patch transfer. In: AAAI. vol.~38, pp. 5070--5078 (2024)

\bibitem{ABS}
Tang, Y., Yu, J., Gai, K., Wang, Y., Hu, Y., Xiong, G., Wu, Q.: Align before search: Aligning ads image to text for accurate cross-modal sponsored search (2023)

\bibitem{I2W}
Tang, Y., Yu, J., Gai, K., Zhuang, J., Xiong, G., Hu, Y., Wu, Q.: Context-i2w: Mapping images to context-dependent words for accurate zero-shot composed image retrieval. In: AAAI. vol.~38, pp. 5180--5188 (2024)

\bibitem{MLC}
Wang, D., Zhang, S.: Unsupervised person re-identification via multi-label classification. In: CVPR. pp. 10978--10987 (2020)

\bibitem{CMPG}
Wang, G., Yang, Y., Zhang, T., Cheng, J., Hou, Z., Tiwari, P., Pandey, H.M.: Cross-modality paired-images generation and augmentation for rgb-infrared person re-identification. Neural Networks  \textbf{128},  294--304 (2020)

\bibitem{OTLA}
Wang, J., Zhang, Z., Chen, M., Zhang, Y., Wang, C., Sheng, B., Qu, Y., Xie, Y.: Optimal transport for label-efficient visible-infrared person re-identification. In: ECCV. pp. 93--109 (2022)

\bibitem{TOP-ReID}
Wang, Y., Liu, X., Zhang, P., Lu, H., Tu, Z., Lu, H.: Top-reid: Multi-spectral object re-identification with token permutation. In: AAAI. vol.~38, pp. 5758--5766 (2024)

\bibitem{SMCL}
Wei, Z., Yang, X., Wang, N., Gao, X.: Syncretic modality collaborative learning for visible infrared person re-identification. In: ICCV. pp. 225--234 (2021)

\bibitem{sysu}
Wu, A., Zheng, W., Yu, H., Gong, S., Lai, J.: Rgb-infrared cross-modality person re-identification. In: ICCV. pp. 5390--5399 (2017)

\bibitem{MCRN}
Wu, Y., Huang, T., Yao, H., Zhang, C., Shao, Y., Han, C., Gao, C., Sang, N.: Multi-centroid representation network for domain adaptive person re-id. In: AAAI. pp. 2750--2758 (2022)

\bibitem{PGMAL}
Wu, Z., Ye, M.: Unsupervised visible-infrared person re-identification via progressive graph matching and alternate learning. In: CVPR. pp. 9548--9558 (2023)

\bibitem{taa}
Yang, B., Chen, J., Ma, X., Ye, M.: Translation, association and augmentation: Learning cross-modality re-identification from single-modality annotation. IEEE Transactions on Image Processing  \textbf{32},  5099--5113 (2023)

\bibitem{yang2023towards}
Yang, B., Chen, J., Ye, M.: Towards grand unified representation learning for unsupervised visible-infrared person re-identification. In: ICCV. pp. 11069--11079 (2023)

\bibitem{SDCL}
Yang, B., Chen, J., Ye, M.: Shallow-deep collaborative learning for unsupervised visible-infrared person re-identification. In: CVPR. pp. 16870--16879 (2024)

\bibitem{ADCA}
Yang, B., Ye, M., Chen, J., Wu, Z.: Augmented dual-contrastive aggregation learning for unsupervised visible-infrared person re-identification. In: ACM MM. pp. 2843--2851 (2022)

\bibitem{DART}
Yang, M., Huang, Z., Hu, P., Li, T., Lv, J., Peng, X.: Learning with twin noisy labels for visible-infrared person re-identification. In: CVPR. pp. 14288--14297 (2022)

\bibitem{CNL}
Yang, M., Huang, Z., Peng, X.: Robust object re-identification with coupled noisy labels. IJCV pp. 1--19 (2024)

\bibitem{CAJ}
Ye, M., Ruan, W., Du, B., Shou, M.Z.: Channel augmented joint learning for visible-infrared recognition. In: ICCV. pp. 13547--13556 (2021)

\bibitem{AGW}
Ye, M., Shen, J., Lin, G., Xiang, T., Shao, L., Hoi, S.C.H.: Deep learning for person re-identification: {A} survey and outlook. {IEEE} Trans. Pattern Anal. Mach. Intell. pp. 2872--2893 (2022)

\bibitem{EP}
Ye, M., Wang, Z., Lan, X., Yuen, P.C.: Visible thermal person re-identification via dual-constrained top-ranking. In: IJCAI. pp. 1092--1099 (2018)

\bibitem{RPL}
Yin, X., Shi, J., Zhang, Y., Lu, Y., Zhang, Z., Xie, Y., Qu, Y.: Robust pseudo-label learning with neighbor relation for unsupervised visible-infrared person re-identification. CoRR  \textbf{abs/2405.05613} (2024)

\bibitem{MEB}
Zhai, Y., Ye, Q., Lu, S., Jia, M., Ji, R., Tian, Y.: Multiple expert brainstorming for domain adaptive person re-identification. In: {ECCV}. vol. 12352, pp. 594--611 (2020)

\bibitem{CCL}
Zhang, G., Zhang, H., Lin, W., Chandran, A.K., Jing, X.: Camera contrast learning for unsupervised person re-identification. {IEEE} Trans. Circuits Syst. Video Technol.  \textbf{33}(8),  4096--4107 (2023)

\bibitem{MT}
Zhang, P., Wang, Y., Liu, Y., Tu, Z., Lu, H.: Magic tokens: Select diverse tokens for multi-modal object re-identification. In: CVPR. pp. 17117--17126 (2024)

\bibitem{FMCNet}
Zhang, Q., Lai, C., Liu, J., Huang, N., Han, J.: Fmcnet: Feature-level modality compensation for visible-infrared person re-identification. In: CVPR. pp. 7339--7348 (2022)

\bibitem{DEEN}
Zhang, Y., Wang, H.: Diverse embedding expansion network and low-light cross-modality benchmark for visible-infrared person re-identification. In: CVPR. pp. 2153--2162 (2023)

\bibitem{MMN}
Zhang, Y., Yan, Y., Lu, Y., Wang, H.: Towards a unified middle modality learning for visible-infrared person re-identification. In: ACM MM. pp. 788--796 (2021)

\bibitem{LAW}
Zhang, Z., Xie, Y., Li, D., Zhang, W., Tian, Q.: Learning to align via wasserstein for person re-identification. IEEE Transactions on Image Processing  \textbf{29},  7104--7116 (2020)

\bibitem{DCMIP}
Zou, C., Chen, Z., Cui, Z., Liu, Y., Zhang, C.: Discrepant and multi-instance proxies for unsupervised person re-identification. In: ICCV. pp. 11058--11068 (2023)

\bibitem{UFineBench}
Zuo, J., Zhou, H., Nie, Y., Zhang, F., Guo, T., Sang, N., Wang, Y., Gao, C.: Ufinebench: Towards text-based person retrieval with ultra-fine granularity. In: CVPR. pp. 22010--22019 (2024)

\end{thebibliography}
\end{document}